\documentclass[conference]{IEEEtran}
\usepackage{cite}
\usepackage{amsmath,amssymb,amsfonts}
\usepackage{graphicx}
\usepackage{textcomp}
\usepackage{dsfont}
\usepackage{xcolor}
\usepackage{algpseudocode}
\def\BibTeX{{\rm B\kern-.05em{\sc i\kern-.025em b}\kern-.08em
    T\kern-.1667em\lower.7ex\hbox{E}\kern-.125emX}}
\begin{document}
\title{Exploring Credibility Scoring Metrics of Perception Systems for Autonomous Driving\\
}

\author{\IEEEauthorblockN{Viren Khandal}
\IEEEauthorblockA{\textit{University of California, Berkeley} \\
Berkeley, California, United States \\
virenkhandal@berkeley.edu}
\and
\IEEEauthorblockN{Arth Vidyarthi}
\IEEEauthorblockA{\textit{University of California, Berkeley} \\
Berkeley, California, United States \\
arthvid@berkeley.edu}
}
\maketitle
\vspace*{-\baselineskip}
\begin{abstract}
% add problem statement, consequences of problem, what is missing from previous work
% below paragraph states solution/contributions and consequences of solution
Autonomous and semi-autonomous vehicles’ perception algorithms can encounter situations with erroneous object detection, such as misclassification of objects on the road, which can lead to safety violations and potentially fatal consequences. While there has been substantial work in the robustness of object detection algorithms and online metric learning, there is little research on benchmarking scoring metrics to determine any possible indicators of potential misclassification. An emphasis is put on exploring the potential of taking these scoring metrics online in order to allow the AV to make perception-based decisions given real-time constraints.
In this work, we explore which, if any, metrics act as online indicators of when perception algorithms and object detectors are failing. Our work provides insight on better design principles and characteristics of online metrics to accurately evaluate the credibility of object detectors. Our approach employs non-adversarial and realistic perturbations to images, on which we evaluate various quantitative metrics. We found that offline metrics can be designed to account for real-world corruptions such as poor weather conditions and that the analysis of such metrics can provide a segue into designing online metrics. This is a clear next step as it can allow for error-free autonomous vehicle perception and safer time-critical and safety-critical decision-making.

% If such a relevant metric, that can indicate, even slightly, that an object may be misclassified and the output of an AV's perception algorithm may be erroneous, the self-driving module can be disengaged and the driver can be brought back into the decision making loop to make any time-critical and/or safety-critical decisions to prevent any safety violations.
\end{abstract}


\section{Introduction}

% \textcolor{red}{YP: I can't be an author on the submission to the ITS workshop, it'd be a conflict of interest. Shouldn't change anything for you though, and you can add my name as an acknowledgement later on. Also, drop the VeHICaL use here, it's not an organization and project names aren't usually mention in the author affiliation information.}

% Key motivation: AV perception can be erroneous and errors can lead to fatal accidents 
% 1. https://www.nytimes.com/2018/03/19/technology/uber-driverless-fatality.html, https://en.wikipedia.org/wiki/Death_of_Elaine_Herzberg
% 2. 

The safe operation of autonomous and semi-autonomous vehicles (AVs) requires accurate predictions from the perception module built into the AV. Modern perception modules rely on object detection models to accurately gather information about the surrounding objects and the road. However, there have been several incidents in which erroneous predictions from these object detection models have led to fatal accidents (Wakabayashi, 2018). 

% discuss how object detectors work and may fail
% discuss YOLO
Modern object detectors such as YOLO and Fast R-CNN rely heavily upon two main tasks: \textbf{object localization}, the process of locating or identifying the presence of an object in an image, and \textbf{image classification}, predicting the type of object and assigning it a label. As such, the YOLO object detector model outputs a label (specifying the object's classification), a bounding box (specifying the object's location), and a confidence score (specifying the model's confidence in the classification and bounding box). Thus, an error in the object detector can arise in one of two possible ways: $(i)$ incorrectly classifying an object and $(ii)$ incorrectly sizing the bounding box. These propagate into the motion planning modules of an AV, and can result in unsafe behaviors form the AV. 

An example of error $(i)$, misclassification of an object, can be seen in Figure 1, in which the pedestrian walking on the side of the road is classified as a cyclist with a very low confidence score of 0.1185. Though the confidence of the model is extremely low, it still chooses to classify this image, which it does so incorrectly. A simple misclassification like this can result in dangerous AV decision making.

On the other hand, an example of error $(ii)$, suboptimal characterization of a bounding box, can be seen in Figure 2, in which the unnecessarily large bounding box of the closest car, hinders the YOLO model's classification of the adjacent motorcycle. Similar to the effects of misclassification in Figure 1, the suboptimal bounding box and its resulting impact on the classification of another object on the road can pose errors in AV decision making as some of the critical on-road objects are now being overlooked.

These are just two of the many examples that motivate the safer operation of AVs through the lens of error-free object detectors and perception algorithms. Improving these perception systems requires identifying (or creating) intrinsic properties or metrics of an image that suggests whether such a misclassification is probable. Furthermore, these metrics must work online when an AV is driving on the road, meaning they cannot be reliant at run-time on pre-labelled data but instead must be computed from intrinsic properties of the image. This can be used to alert the driver to a potential failure of the autonomous system and take control if needed.

% new paragraph about credibility vs robustness
We choose to to further study the credibility of these object detectors rather than the robustness. In the context of reliable autonomous driving, we choose to dive deeper into credibility rather than robustness because credibility is used to determine the reliability of the AV's judgement, which, if low, can be used to transfer driving control from the AV to the human. Clearly defined, credibility and robustness differ in that credibility measures the effectiveness of object detectors on non-adversarial perturbations, whereas robustness primarily deals with adversarial perturbations. 

% Include parts of this paragraph somewhere (especially second sentence)
In this paper, we discuss \textit{various metrics that act as indicators of when perception algorithms might fail to correctly classify on-road objects}. While these metrics are not directly capable of being deployed online, they still provide a valuable foundation from which online metrics can be designed that incorporate prior knowledge about a classifier's offline robustness. We hope to use these empirical results to better understand when the self-driving component can be disengaged and control can be shifted back to the driver to make proper decisions.

In the rest of this paper, we discuss some related/prior works that outline metrics for analyzing robustness, we then discuss our core approach and experimental setup, and conclude with qualitative and quantitative core results/takeaways.

\begin{figure}[h]
    \centering
    \includegraphics[width=.3\textwidth]{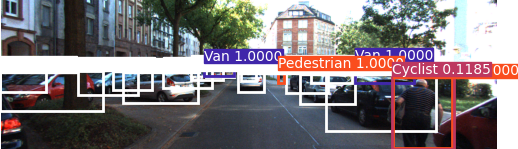}
    \caption{On the right side of the image, a pedestrian who is currently walking to the side of the road is classified as a cyclist, rather than a pedestrian.}
\end{figure}

\begin{figure}[h]
    \centering
    \includegraphics[width=.3\textwidth]{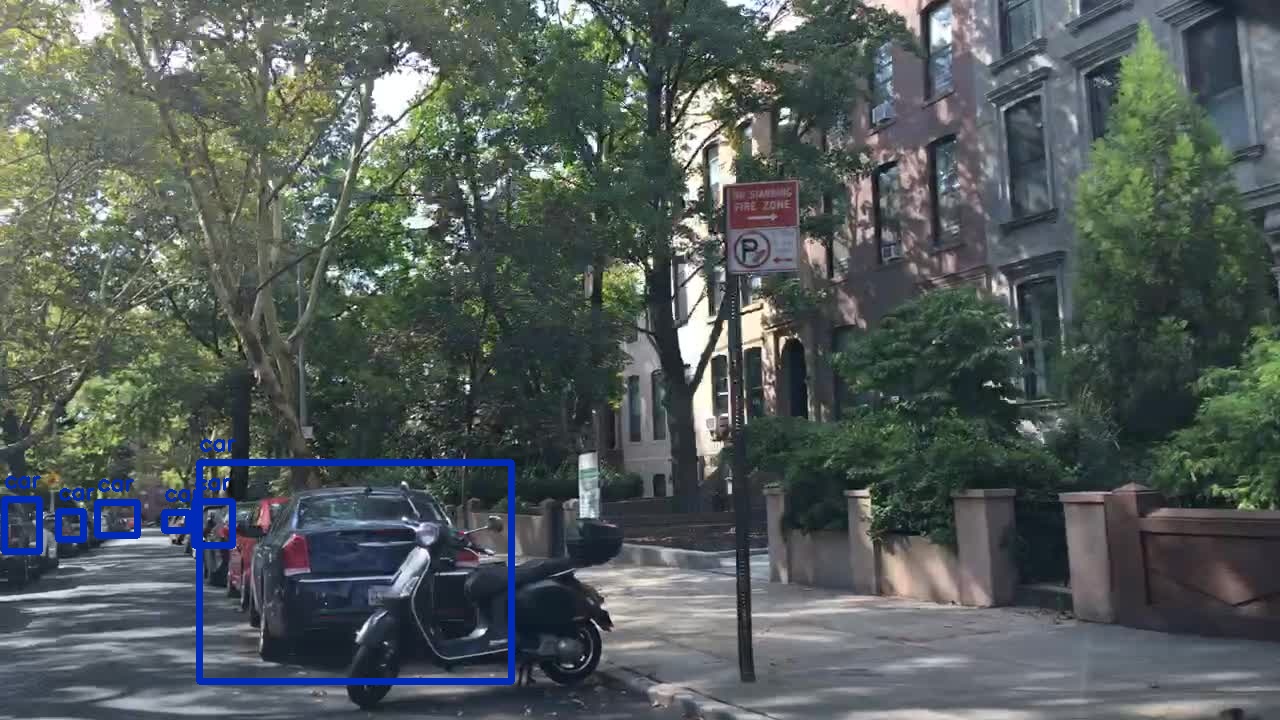}
    \caption{The decreased confidence in the prediction of the closest car, leads to the model predicting a larger bounding box, which then impairs it from also predicting the motorcycle positioned directly behind it.}
\end{figure}

% \textcolor{red}{YP: But that's not what you're doing in this paper. The VeHICaL setup is also not a known thing, so let's not mention it. Treat this as a standalone work. Motivation should be a part of the introduction, doesn't need a new section/subsection or even heading. Remember, the motivation is that AV perception can be erroneous and the errors can be fatal (cite a famous accident or two here). All you want to do in this work is explore what quantities/metrics act as indicators (online) of when these perception algorithms are failing. The bigger picture is that is this information can be used to disengage self-driving and bring the driver back into the decision making loop. Don't complicate things by talking about a perception hand-off etc.}

\section{Related Work}
Hendrycks et al., [2] focused on benchmarking image classifier robustness. It explored the use of both adversarial perturbations along with visual corruptions on the ImageNet dataset in order to score classifiers on their ability to correctly identify classes within these augmented images. The paper focused on common visual corruptions and also showed how classifiers can be trained to become more robust to such artifacts. In doing so the authors established several unique metrics (\textbf{mean Corruption Error} and \textbf{Flip Probability}), some of which we used as inspiration for our own custom metrics. We further extended their work on classifier robustness to scoring object detector robustness.

Further, Xue et al., 2021 [3] proposes adversarial attacks on object detectors based on natural perturbations. The authors introduce the concept of \textbf{Real-World Perturbation Score (RPS)} which quantifies the similarity of adversarial attacks to real noises in the physical world. While their work focuses on adversarial attack methods to understand robustness of various object detectors (YOLO, Fast R-CNN, and SSD), we extend their work on real-world adversarial perturbations by focusing on understanding the intrinsic properties of an image, measured with non-adversarial metrics, that may suggest misclassification. We maintain their ideas of \textit{RPS} by using real-world perturbations, including Gaussian Noise/Blur and visual corruptions (fog, sunflare, and snow).

On the other hand, Zhao et al. [4] distinguish the types of adversarial attacks into Hiding Attacks, which make the object detector fail to recognize the object, and Appearing Attacks, which make the object detector mis-recognize the object. Our previous discussion of the two possibilities of error in object detection, either in object localization or image classification, reflects a parallel structure to the study mentioned in Zhao et al. Consequently, the paper suggests \textbf{Enhanced Realistic Constraints Generation (ERGs)}, which aim to use context from images such as the background to build an adversarial attack. Though this concept is suggested, the study mainly injects a variety of random patterns and shapes into images, which are unlikely to occur in the physical world. We expand on their study by constructing our perturbations to be non-adversarial, all the while using the paper’s suggestions on \textit{ERGs} to make our corruptions to the images more natural and realistic.

Other work by Durall et al. [5] looks at images converted to the frequency domain, specifically GAN generated deepfakes, in order to determine their credibility. The idea here is to achieve a metric for credibility that does not depend on labels associated with the data but rather on image intrinsics itself. Our study extends this thesis of image intrinsics being used as a credibility measure to object detectors, in the hopes that the corruption metrics we analyze can contribute to the development on label-independent online methods for scoring these models.

% \textcolor{red}{Still needs more references. Also, make table captions more descriptive if there's space. }
% \textcolor{red}{This needs more references, and the citation form is not correct. e.g., Hendrycks et al.,\cite{b1}. References don't need to be bolded either. In general, you should have at least 10-odd references overall in the paper, since robustness of perception systems is an established area (although your angle of attack is new, but still). Feel free to be a bit reductive while referring to papers here, in the interest of space.}

\section{Approach}
% \textcolor{red}{1. To start off, clearly define (aka restate) the problem that you're trying to solve. Use a bold text heading (see below) and define it in no more than 3 lines. 2. After that, talk about how you achieve it with your method, 3. Then get into the details of your approach. You can use bold text headings for the 'problem statement', and 'Proposed approach'.}

\noindent\textbf{Problem Statement:} Offline metrics for scoring the reliability of perception systems do not take into account the distribution shift that occurs at test time. This can lead to poor predictions that pose a serious threat to driver safety in semi-autonomous vehicles.

\noindent\textbf{Proposed Approach:} Corrupt images with realistic artifacts found during on-road deployment, calculate series of metrics, and analyze if any of the calculated metrics serve as indicators of possible online misclassification. Use these metrics as a base to then design intelligent online scoring metrics.
\\

Our first step in studying the credibility of these object detector models was to create a main method to adjust the original image and determine any intrinsic properties that change within the images that may imply misclassification. 

% \textcolor{red}{What previously computer metric? This is hard not clear.}

% \textcolor{red}{Be more direct in the writing, e.g., In our proposed approach, we use non-adversarial corruptions ($<$what kind? additive to the image?$>$) ....}

We chose to abstain from using adversarial perturbations which while known to confuse deep neural networks (Goodfellow et al., 2014)[6], are not good indicators of whether or not a network is learning robust latent space representations of the input classes. 

In our proposed approach, we used several types of \textbf{non-adversarial corruptions}. Image classifiers are known to achieve subhuman accuracy when images are perturbed with Gaussian artifacts (Dodge et al., 2017) [7]. To test this, one set of experiments were hence with normally distributed artifacts in the form of Gaussian noise and Gaussian blur, both of which add some probability-based input to marginally corrupt the image. 

The second tactic we used was corruptions that would reflect realistic road conditions that an AV might encounter, such as extreme glare from surrounding vehicles or bad weather like fog and rain. 

% In order to create a concrete framework for the outlined method, we decided to use corruptions that were not adversarial in nature (what one would expect in a malicious attack on a classifier) but more similar to realistic road conditions that an AV might encounter that could lead to it misclassifying objects on the road.

One advantage of this approach is that it could be used in an online method to make any relevant perception-based decisions given real time constraints. Furthermore, this approach uses perturbations that can be observed in the real world, thus maintaining a high real-world perturbation score and ecological validity (Xue et al., 2021). 

% New paragraph about choosing not to train on perturbed/ corrupted images.
One important factor that's important to highlight in our methodology is that we purposely chose not to train our detector on corrupted images. The intuition behind this decision is as follows. Consider the case of an arbitrary detector whose training data we have no prior knowledge about. Such a detector, regardless of its level of robustness, would be expected to perform poorly on corrupt samples \textbf{relative} to more natural data. The issue with using a detector trained on a corrupt dataset is that it becomes difficult to benchmark performance across a variety of detectors. Our choice of metrics have been designed to be benchmarked using a clean detector so that it can serve as a baseline for more robust detectors that share the same backbone but are trained on corrupt data samples.

% \textcolor{red}{More importantly, is this the only approach you both followed? I suspect there's more going methodology in the results and finding section than there is in this section. }

% \textcolor{red}{cite something to back this up}.  \textcolor{red}{Perhaps flip the order of these benefits.}

% explain how perturbations may occur in real scenario
% if image is blurred or data is somehow corrupted
\section{Results and Findings}
% \textcolor{red}{You need a subsection to first explain the setup (detectors, datasets, labels etc.) and what your empirical studies are trying to achieve. That will help you make the following text more crisp as well, e.g., you no longer need statements like 'such as YOLO'.}
% \subsection{Setup}
For our project setup, we used a YOLOv3 object detector (Redmon et al., 2018) [8] for all the trials carried out in this paper. We used two versions of the detector: one trained on the Berkeley Deep Drive Dataset (Yu et al., 2020) [9] and the other trained on the KITTI Vision Benchmark Suite (Geiger et al., 2012)[10]. We used a combination of noising and blurring perturbations to effectively model real-world scenarios, such as fog, light glare, and snow.

% We chose to use YOLO-v3 over newer versions such as YOLO-v4 and v5 primarily due to the speed up in training time. It is also important to note that our focus is not on architecture-centric but data-centric, looking at how changing the quality of data being passed into an arbitrary detector can impact its performance relative to a clean baseline. This means that while alternate approaches using newer architectures may achieve better absolute results on our metrics, the relationship between the metrics itself depending on the quality of data (clean vs corrupted) should remain relatively stable. 

\subsection{Average Label Confidence}\label{AA}
\noindent\textbf{Key Results:} Average label confidences of the YOLOv3 model without any perturbations were \textbf{0.953}, whereas with both Gaussian Noise and Gaussian Blur perturbations the average label confidence dropped nearly \textbf{10\%} to \textbf{.858}. Blurring perturbations had a significantly larger impact than noising perturbations, which shows that realistic blurry conditions, such as fog, can decrease the model's confidence levels enough to cause errors in perception.
\\
When object detector models, such as YOLO, make predictions of objects on the road, they supply 3 main components: a label which corresponds the category of the detected object, a bounding box with proper coordinates, and a confidence metric. This confidence metric, as seen in its name, gives an estimate for the confidence of the prediction, which is based on a cross-entropy loss function.

We leveraged this confidence measure to gauge how the object detector model behaves when supplied with corrupted/partially perturbed images, rather than the normal images it was trained on. The below Table I shows the results from our average confidence studies.

% \textcolor{red}{Are these really 'metrics' or just 'measures'? Check the definition of these two terms and see where your functions fall.}

\begin{table}[htpb]
\begin{center}
\begin{tabular}{|c|c|c|}
  \hline
  & \textbf{No Gaussian Blur} & \textbf{Gaussian Blur} \\[2pt]
  \hline
  \textbf{No Gaussian Noise} & 0.953 & 0.887 \\[2pt]
  \hline
  \textbf{Gaussian Noise} & 0.922 & 0.858 \\[2pt]
  \hline
\end{tabular}
\end{center}
\caption{Confidence Scores based on various perturbations}
\end{table}
\vspace*{-\baselineskip}
As can be seen in the table, the confidence of the model's predictions decreases by nearly 10\% with both Gaussian noise and Gaussian blur perturbations. This drop in confidence is a crucial metric which can imply when a model's predictions are credible, as it can be used in an online scenario by comparing it to a threshold. Comparing the confidence in our model's predictions to a higher threshold can allow the AV's perception module to make more accurate predictions, thus decreasing the misclassification error \textit{(See subsection C)}.

In fact, a direct result of the drop in confidence can expand the size of a bounding box around some object, which may lead to not detecting other objects in close vicinity. For example, Figure 2 shows how the drop in confidence of one object leads to a much larger bounding box than needed, thus not identifying the motorcycle behind it.

% \begin{figure}[h]
%     \centering
%     \includegraphics[width=.5\textwidth]{large_bounding_box.jpg}
%     \caption{The decreased confidence in the prediction of the closest car, leads to the model predicting a larger bounding box, which then impairs it from also predicting the motorcycle positioned directly behind it. \textcolor{red}{This example is good! Possibly make this figure 1 and use it to motivate the idea(s)?}}
% \end{figure}

% consider adding example image that shows confidence level dropping and bounding box getting larger

\subsection{Mean Corruption mAP}
The correctness of predictions made by image classifiers is often heavily dependent on the quality of input images passed in. Real-world input data streams that are passed into the perception modules of self-driving vehicles are not guaranteed to be drawn from the same distribution as the training data. Online streams are much more prone to visual corruptions such as fog, excessive brightness, and blur.

We benchmark the robustness of our YOLO detector using a metric inspired by Mean Corruption Error (mCE) introduced by Hendrycks et al. mCE acts as an aggregate performance measure by comparing how a classifier's error rate changes across different visual corruptions of varying severities. Our approach works on an almost identical principle to mCE but instead we computed the Mean Corruption mAP (MCmAP) since our focus is on a detection and classification task, requiring a different base metric of accuracy. 

We first defined our classifier $f$ as a YOLOv3 model trained on the KITTI 2D Object Detection dataset. The classifier was trained on 7481 images with 7 possible classes. We further define a set of corruptions $C$ containing the following real-world visual artifacts: $C = \{snow, sunflares, fog\}$. We also define three levels of intensity for each: $I = \{low, medium, high \}$. We then compute the following score for each corruption $c \in C$:

\begin{equation}
CmAP_{i}^{c} = \sum_{i=low}^{high} mAP_{i, c}
\end{equation}

The computation is explained as follows: we first computed the mean average precision over all 7 classes for a given corruption $c$ and a given intensity $i$. We then summed over all possible intensities to arrive at the corruption mAP or CmAP for the given corruption.

Finally we compute the MCmAP by averaging the CmAP values over all three corruption classes:
\begin{equation}
MCmAP = \frac{\sum_{c} CmAP_{c}}{3}
\end{equation}
The process of computing this metric allowed us to gauge several interesting characteristics of our classifier.
\\
\subsubsection{Class Robustness}
We were able to see how the average precision (AP) on certain classes was more robust to visual corruptions than others. This was useful in helping analyze whether certain classes should warrant lower confidence scores given their predisposition to be incorrectly classified under certain weather conditions. The tables below showcase the AP for each of the 7 classes for each of the 3 corruptions (along with their intensities):

\begin{table}[htpb]
\begin{center}
\begin{tabular}{|c|c|c|c|}
  \hline
  & \textbf{Low} & \textbf{Medium} & \textbf {High} \\ \textbf {Difference} & & &  \\
  \hline
  \textbf{Pedestrian} & 0.856 & 0.631 & 0.406 \\ 0.450 & & & \\
  \hline
  \textbf{Cyclist} & 0.809 & 0.449 & 0.140 \\ \textbf{0.669}  & & & \\
  \hline
  \textbf{Car} & 0.975 & 0.792 & 0.347 \\ 0.628  & & & \\
  \hline
  \textbf{Van} & 0.685 & 0.385 & 0.212 \\ 0.473  & & & \\
  \hline
  \textbf{Miscellaneous} & 0.640 & 0.125 & 0.013 \\ 0.627 & & &  \\
  \hline
  \textbf{Truck} & 0.780 & 0.217 & 0.048 \\ \textbf{0.732} & & & \\
  \hline
  \textbf{Person Sitting} & 0.882 & 0.283 & 0.085 \\  \textbf{0.797} & & &  \\
  \hline
  \textbf{Tram} & 0.717 & 0.244 & 0.070 \\ 0.647 & & &  \\
  \hline
\end{tabular}
\end{center}
\caption{Average precision with fog applied}
\end{table}
\begin{table}[htpb]
\begin{center}
\begin{tabular}{|c|c|c|c|}
  \hline
  & \textbf{Low} & \textbf{Medium} & \textbf {High} \\ \textbf {Difference} & & & \\
  \hline
  \textbf{Pedestrian} & 0.729 & 0.570 & 0.390 \\ 0.339  & & & \\
  \hline
  \textbf{Cyclist} & 0.789 & 0.610 & 0.380 \\ 0.409  & & & \\
  \hline
  \textbf{Car} & 0.761 & 0.398 & 0.193 \\ \textbf{0.568} & & &  \\
  \hline
  \textbf{Van} & 0.709 & 0.569 & 0.351 \\ 0.358  & & & \\
  \hline
  \textbf{Miscellaneous} & 0.615 & 0.517 & 0.152 \\ \textbf{0.463}  & & & \\
  \hline
  \textbf{Truck} & 0.651 & 0.545 & 0.332 \\ 0.319  & & & \\
  \hline
  \textbf{Person Sitting} & 0.715 & 0.431 & 0.210 \\ \textbf{0.505}  & & & \\
  \hline
  \textbf{Tram} & 0.729 & 0.557 & 0.321 \\ 0.408 & & & \\
  \hline
\end{tabular}
\end{center}
\caption{Average precision with sunflares applied}
\end{table}
\begin{table}[htpb]
\begin{center}
\begin{tabular}{|c|c|c|c|}
  \hline
  & \textbf{Low} & \textbf{Medium} & \textbf {High} \\ \textbf {Difference}\\
  \hline
  \textbf{Pedestrian} & 0.909 & 0.909 & 0.773 \\ 0.136  & & & \\
  \hline
  \textbf{Cyclist} & 0.920 & 0.920 & 0.685 \\ 0.235  & & & \\
  \hline
  \textbf{Car} & 1.0 & 1.0 & 0.700 \\ 0.300  & & & \\
  \hline
  \textbf{Van} & 0.854 & 0.854 & 0.619 \\ 0.235  & & & \\
  \hline
  \textbf{Miscellaneous} & 0.781 & 0.781 & 0.344 \\ \textbf{0.437}  & & & \\
  \hline
  \textbf{Truck} & 0.933 & 0.933 & 0.614 \\ 0.319  & & & \\
  \hline
  \textbf{Person Sitting} & 0.953 & 0.953 & 0.612 \\ \textbf{0.341}  & & & \\
  \hline
  \textbf{Tram} & 0.767 & 0.767 & 0.410 \\ \textbf{0.357} & & &  \\
  \hline
\end{tabular}
\end{center}
\caption{Average precision with snow applied}
\end{table}

In each of the tables, the difference column indicates the change in average precision for that particular class as the intensity of the corruption was adjusted from low to high. The differences highlighted in bold are the top three largest changes for each corruption.  

For each of the corruptions, the class \textbf{Person Sitting} was consistently in the top 3 classes most affected  by artifacts. However we didn't find this to be a surprising result since the prior probability of this class occurring in the distribution of images selected for inference was marginally lower than the other classes (the only instances are generally people in bus stands). 

\noindent\textbf{Key Results:} The sunflare corruption was applied to mimic the glares that have caused many perception modules to malfunction. Alarmingly, it was the \textbf{Car} class that showcased the largest change in AP as the intensity of the glare was modulated to high: a drop of close to \textbf{0.57}. Note that the use of AP here provides a deeper insight than just classification accuracy: not only is this indicative of poor classification but also that the predicted bounding boxes are significantly worse at localizing cars under weather conditions where glare is a problem. This implies that the confidence underlying these predictions must be discounted under certain external conditions and further work must be done to determine the precise mechanism for doing so. 

\subsubsection{mAP Across Corruptions}
 
We also analyzed how the mAP shifted across different corruption intensities. The results can be found in the table below:  
\\
\begin{table}[htpb]
\begin{center}
\begin{tabular}{|c|c|c|c|}
  \hline
  & \textbf{Low} & \textbf{Medium} & \textbf {High} \\
  \hline
  \textbf{Fog} & 0.793 & 0.391 & 0.165 \\
  \hline
  \textbf{Sunflare} & 0.712 & 0.524 & 0.291 \\
  \hline
  \textbf{Snow} & 0.890 & 0.890 & 0.595 \\
  \hline
\end{tabular}
\end{center}
\caption{mAP across different corruptions}
\end{table}\\
\noindent\textbf{Key Results:} The addition of fog caused the most significant drop in mAP while the addition of snow did not seem to have as adverse of an effect. We attributed this to the mechanism with which both artifacts were applied: the snow augmentation imputed randomly selected pixels with (255, 255, 255) RGB values, effectively whitening them out. Adding fog applied a blur to all pixels evenly (determined by a pre-selected blurring coefficient). This likely lead to edges and areas of high contrast being altered, thus adversely impacting the ability of a detector (particularly a one-stage mechanism such as YOLO).

\subsubsection{Mean Corruption Mean Average Precision (MCmAP}

We finally computed an aggregate performance score for our classifier using the following individual CmAP scores:
\begin{table}[htpb]
\begin{center}
\begin{tabular}{|c|c|}
  \hline
  & \textbf{cMap Scores}  \\
  \hline
  \textbf{Fog} & 1.349 \\
  \hline
  \textbf{Sunflare} & 1.528 \\
  \hline
  \textbf{Snow} & 2.373 \\
  \hline
\end{tabular}
\end{center}
\caption{CmAP across different corruptions}
\end{table}

\noindent\textbf{Key Results:} Using these and our formula from the start of this subsection, we have our final MCmAP score for this classifier as: \textbf{1.75}.

It is expected that as a classifier becomes more robust to corruptions, its MCmAP score will increase in value. While this remains an offline metric, knowledge of an MCmAP score along with its various subscores for each corruption, intensity, and class, can provide valuable insight into shaping online metrics that account for real-time weather conditions. 

\subsection{Average Misclassification Error}
\noindent\textbf{Key Results:} Misclassification error is a crucial metric, as it is a direct measurement of how poorly the image classification component of the YOLOv3 model works. Without any perturbations, the model has a misclassification error of \textbf{0.141}, which  means it misclassifies objects in an image nearly \textbf{14\%} of times. Moreover, evaluating our model on perturbed images nearly doubles the misclassification error, approximately \textbf{0.27}. Thus, any online metric that aims to serve as an indicator of erroneous object detection, must take into consideration the intrinsic properties that cause this misclassification in the first place.
\\
One of the main characteristics we wanted to investigate was misclassification error, or how often does the YOLO object detection model classify some object incorrectly.

\begin{equation}
Misclassifcation\;Error = \frac{FP + FN}{TP + TN + FP + FN}
\end{equation}

Here FP and FN represent False Positives and False Negatives, respectively, where their sum represents the number of incorrectly predicted outcomes. The denominator of Equation (3) is the sum of TP (True Positives), TN (True Negatives), FP (False Positives), and FN (False Negatives), which represents the total number of predictions.

Running this metric on our model, we discovered the average misclassification error to be \textbf{0.141} without any perturbations, which translates to the YOLO model incorrectly predicting a class nearly 14\% of times. This is a significantly higher number than expected, as an industry-leading object detector such as YOLO would hope to be more precise. In fact, this 0.141 misclassification error also implies the precision of the YOLO model is approximately \textbf{86\%}, meaning there is substantial improvement needed in order to deploy these models safely.

With added perturbations, the misclassification error, as expected, nearly doubled. The detailed results of this can be seen in the table below.

\begin{table}[htpb]
\begin{center}
\begin{tabular}{|c|c|c|}
  \hline
   & \textbf{No Gaussian Blur} & \textbf{Gaussian Blur} \\[2pt]
  \hline
  \textbf{No Gaussian Noise} & 0.141 & 0.265 \\[2pt]
  \hline
  \textbf{Gaussian Noise} & 0.252 & 0.269 \\[2pt]
  \hline
\end{tabular}
\end{center}
\caption{\scriptsize{Misclassification Error based on Gaussian Noise and Blur}}
\end{table}
\vspace*{-2\baselineskip}
\subsection{Flip Probability} 
\noindent\textbf{Key Results:} Calculating Flip Probability, a measurement of how likely a model is to change its prediction based on some perturbations, gave us deeper insights on how credible the predictions of our YOLOv3 were. Inputting the non-perturbed images gave a flip probability of \textbf{0}, meaning the model was deterministic and would be consistent with its predictions. However, once we layered on some noise and blur through gaussian perturbations, the flip probability expanded to \textbf{0.1842}, meaning nearly \textbf{18\%} of the time our model would flip/alter class labels. Again, much like confidence levels, flip probability was influenced more by blurring effects, rather than noise.
\\
The next metric we explored to determine the credibility of the YOLO Object Detector was Flip Probability (Hendrycks et al.). Grounded heavily in misclassification error, the Flip Probability of a network measures the probability that a network's output is flipped or changed based on a series of perturbations.

We went about calculating Flip Probability in 2 phases. First, we created a sequence of $m$ perturbed versions of $n$ images based on two main perturbations: gaussian noise and gaussian blur. 
These perturbations were performed at various degrees, ranging from 0 to 5, to adjust the weight of the perturbations on each image. Then, we performed a misclassification error calculation, which was averaged at the end in order to give a Flip Probability score for the entire YOLO model. The following equation (4) outlines the formula used to calculate Flip Probability, where each \(x^{(i)}_{j}\) represents the \(j^{th}\) perturbed copy of the \(i^{th}\) image.

% insert equation
\begin{equation}
Flip\;Probability=\frac{1}{m(n-1)}\sum_{i=1}^{m}\sum_{j=2}^{n} \mathds{1}(f(x^{(i)}_{j}) \neq f(x^{(i)}_{j-1}))
\end{equation}

Running this Flip Probability test, the Flip Probabilities of the YOLOv3 model are shown in Table VIII, which implies that slight perturbations in the images passed into an object detector model can cause significant misclassifications in the predictions (Fig 3).
\\
\begin{table}[htpb]
\begin{center}
\begin{tabular}{|c|c|c|}
  \hline
   & \textbf{No Gaussian Blur} & \textbf{Gaussian Blur} \\[2pt]
  \hline
  \textbf{No Gaussian Noise} & 0 & 0.1835 \\[2pt]
  \hline
  \textbf{Gaussian Noise} & 0.0039 & 0.1842 \\[2pt]
  \hline
\end{tabular}
\end{center}
\caption{Flip Probabilities based on Gaussian Noise and Gaussian Blur perturbations}
\end{table}
\begin{figure}[]
    \centering
    \includegraphics[width=.3\textwidth]{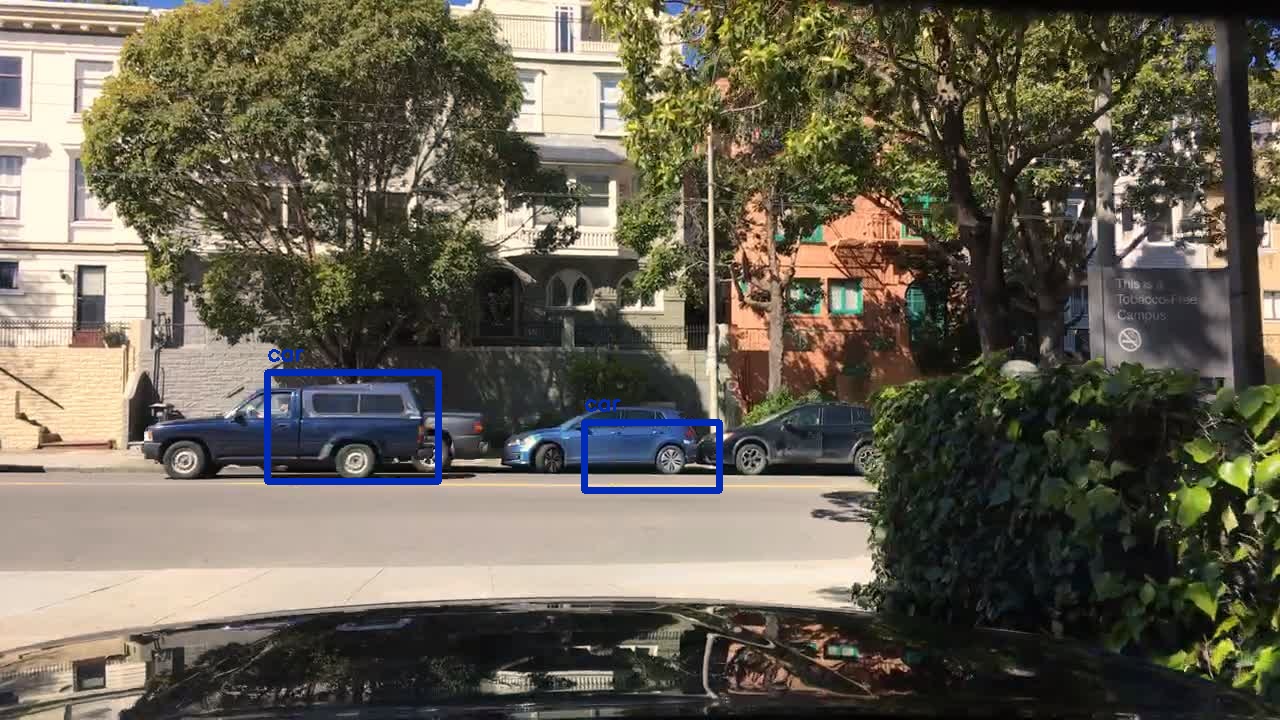}
    
    \vspace{5pt}
    
    \centering
    \includegraphics[width=.3\textwidth]{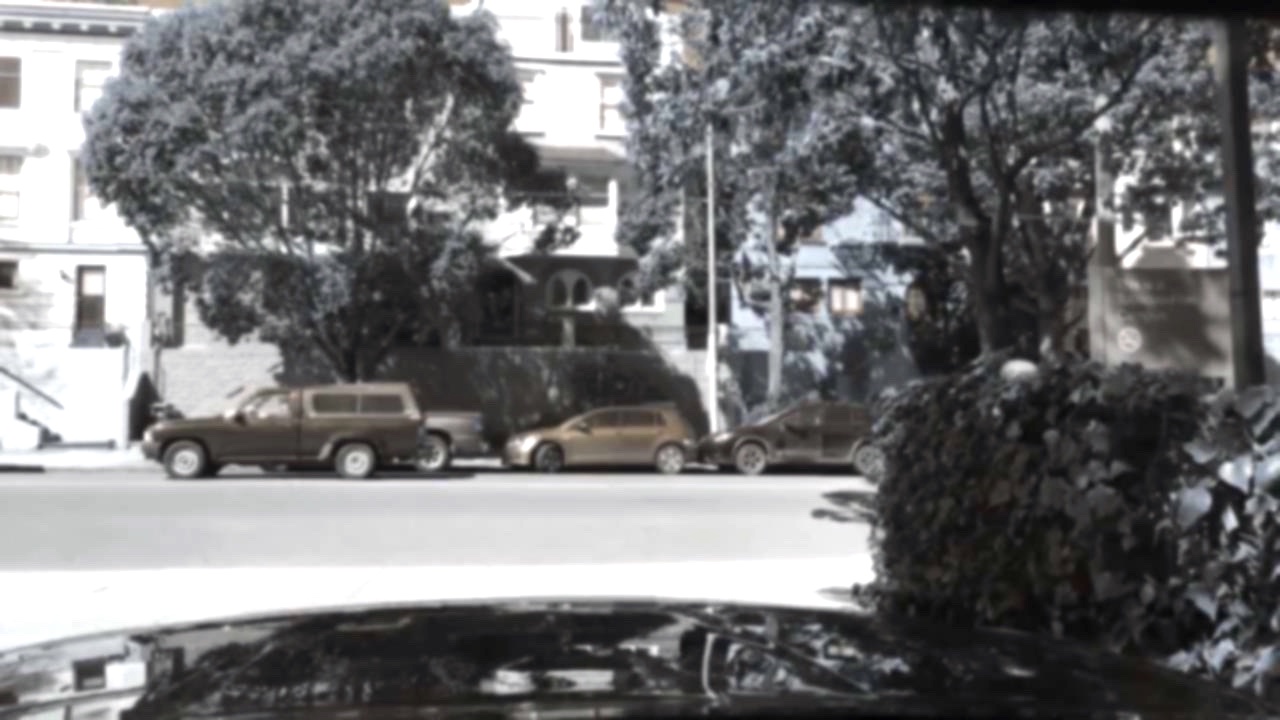}
    \caption{When a corrupted/perturbed image (bottom) is passed into the model, it is unable to properly classify the two main cars in the image, while it does so properly when the clean/non-perturbed image (top) is passed in.}
\end{figure}
% keep qualitative section only if necessary, else combine with other sections and show relevancy (FFT is mainly qualitative - arth)
% \subsection{Qualitative Results}

% discuss qualitative results (arth and viren)
\vspace*{-2\baselineskip}
\section{Conclusion}

The metrics discussed above extend earlier work on benchmarking the robustness of image classifiers to object detectors. This is a crucial extension since AV safety is a dual task of classifying objects correctly along with localizing them with a high degree of precision. This task is made harder when we consider how easy it is for the input data stream to an AVs perception module to be corrupted in a way that shifts its distribution too far from the training data to a point where the module's predictions are too volatile to be considered safe. 

Benchmarking on YOLO allowed us to see how a state-of-the-art one-stage object detector might perform in an online setting. On clean data our detector achieved a Top-1 Accuracy of \textbf{86\%} indicating strong offline reliability. To simulate online performance on unseen data, we chose to create corruptions in the dataset that mimic real-world noise. This was done through both Gaussian perturbations along with simulated weather conditions. We observed that certain classes that have a lower prior distribution in the training data tend to suffer the most when augmented with any such visual corruption. We also noticed that \textbf{glare} remains a large problem specifically on cars that causes a large drop in mAP. 
While these metrics are not designed to be run online, they do give us insight into how to better design online metrics that can leverage this base. For instance, knowing which classes have lower prior probabilities of occurring in our training data can allow us to enforce a lower online entropy constraint on the softmax predictions for images containing said classes. Furthermore we could also account for real-time weather conditions such as during times of fog or snow. Poor weather conditions could automatically enforce harsher online constraints on the minimum confidence threshold we require on a prediction to allow the AV to retain control of the driving task. 

% added another improvement paragraph about further investigation of yolov4 and yolov5
% Another step of improvement we hope to take is evaluating these metrics on YOLOv4 and YOLOv5, both of which have an approximately 10\% increase in accuracy and 12\% increase in frame rate.

Developing this base into completely online metrics remains a task for future exploration, but one that is motivated by the knowledge that developing a fuller understanding of these metrics is crucial in making the future of AVs a safe and feasible reality.

% If such a relevant metric, that can indicate, even slightly, that an object may be misclassified and the output of an AV's perception algorithm may be erroneous, the self-driving module can be disengaged and the driver can be brought back into the decision making loop to make any time-critical and/or safety-critical decisions to prevent any safety violations.
% discuss relevancy to AV security (data corruption)

\section{Acknowledgements}
We would like to thank Professor Yash Vardhan Pant for meaningful discussions and helpful suggestions during the course of this work. We would also like to thank the COMSNETS Intelligent Transportation Systems Workshop 2022 anonymous reviewers for their helpful comments.
\\

% fix bibliography by citing properly

\vspace{12pt}


\begin{thebibliography}{00}
\bibitem{b1} Daisuke Wakabayashi, “Self-Driving Uber Car Kills Pedestrian in Arizona, Where Robots Roam", in New York Times, 29 Mar 2018
\vspace{5pt}
\bibitem{b2} Dan Hendrycks and Thomas Dietterich, “Benchmarking Neural Network Robustness to Common Corruptions and Perturbations,” in International Conference of Learning Representations, ICLR 2019, 28 Mar 2019
\vspace{5pt}
\bibitem{b3} Mingfu Xue, Chengxiang Yuan, Can He, Jian Wang, and Weiqiang Liu, “Naturalae: Natural and robust physical adversarial examples for object detectors," in Journal of Information Security and Applications, 2021.
\vspace{5pt}
\bibitem{b4} Yue  Zhao, Hong Zhu, Ruigang Liang, Qintao Shen, Shengzhi  Zhang,and Kai Chen, “Seeing isn’t believing: Towards more robust adversarial attack against real world object detectors,” in Proceedings  of the 2019 ACM SIGSAC Conference on Computer and Communications Security,New York, NY, USA, 2019, CCS ’19, p. 1989–2004, Association for Computing Machinery
\vspace{5pt}
\bibitem{b5} Ricard Durall, Margret Keuper, Franz-Josef Pfreundt, Janis Keuper, "Unmasking DeepFakes with simple Features," 4 March 2020
\vspace{5pt}
\bibitem{b6} Ian J. Goodfellow, Jean Pouget-Abadie, Mehdi Mirza, Bing Xu, David Warde-Farley, Sherjil Ozair, Aaron Courville, and Yoshua Bengio, “Generative Adversarial Networks,” 10 June 2014
\vspace{5pt}
\bibitem{b7} Samuel F. Dodge and Lina J. Karam, “A study and comparison of human and deep learning recognition performance under visual distortions", in Computing Research Repository, CORR, 2017.
\vspace{5pt}
\bibitem{b8} Joseph Redmon and Ali Farhadi, “YOLOv3: An Incremental Improvement”, in arXiv Preprint, 8 Apr 2018
\vspace{5pt}
\bibitem{b9} Fisher Yu, Haofeng Chen, Xin Wang, Wenqi Xian, Yingying Chen, Fangchen Liu, Vashisht Madhavan, and Trevor Darrell, “BDD100K: A Diverse Driving Dataset for Heterogeneous Multitask Learning," in The IEEE Conference on Computer Vision and Pattern Recognition (CVPR), 8 Apr 2020
\vspace{5pt}
\bibitem{b10} Andreas Geiger, Philip Lenz, and Raquel Urtasun, “Are we ready for Autonomous Driving? The KITTI Vision Benchmark Suite”, in The IEEE International Conference on Robotics and Automation, ICRA 2012, 2012
\vspace{5pt}
\bibitem{b11} Ramya. A, Venkateswara Gupta Pola, Dr. Amrutham Bhavya Vaishnavi, Sai Suraj Karra, "Comparison of YOLOv3, YOLOv4 and YOLOv5 Performance for
Detection of Blood Cells", April 2021

\end{thebibliography}
\end{document}